\documentclass[10pt,twocolumn,letterpaper]{article}

\usepackage{3dv}
\usepackage{times}
\usepackage{epsfig}
\usepackage{graphicx}
\usepackage{amsmath}
\usepackage{amssymb}
\usepackage{textcomp}
\usepackage{authblk}
\usepackage{booktabs}
\usepackage[table,xcdraw]{xcolor}

\newcommand{\degree}[1]{${#1}^o$}
\def\360{\degree{360}}

\usepackage[pagebackref=true,breaklinks=true,letterpaper=true,colorlinks,bookmarks=false]{hyperref}

\threedvfinalcopy 

\def\threedvPaperID{223} 

\begin{document}

\title{Spherical View Synthesis for Self-Supervised \360 Depth Estimation}

\author[1, 2]{Nikolaos Zioulis}
\author[1]{Antonis Karakottas}
\author[1]{Dimitrios Zarpalas}
\author[2]{Federico Alvarez}
\author[1]{Petros Daras}

\affil[1]{\small Information Technologies Institute (ITI), Centre for Research and Technology Hellas (CERTH), Greece}
\affil[2]{\small Signals, Systems and Radiocommunications Department (SSRD), Universidad Polit\'{e}cnica de Madrid (UPM), Madrid, Spain}

\maketitle

\begin{abstract}
   Learning based approaches for depth perception are limited by the availability of clean training data. 
    This has led to the utilization of view synthesis as an indirect objective for learning depth estimation using efficient data acquisition procedures. 
    Nonetheless, most research focuses on pinhole based monocular vision, with scarce works presenting results for omnidirectional input.
    In this work, we explore spherical view synthesis for learning monocular \360 depth in a self-supervised manner and demonstrate its feasibility.
    Under a purely geometrically derived formulation we present results for horizontal and vertical baselines, as well as for the trinocular case.
    Further, we show how to better exploit the expressiveness of traditional CNNs when applied to the equirectangular domain in an efficient manner.
    Finally, given the availability of ground truth depth data, our work is uniquely positioned to compare view synthesis against direct supervision in a consistent and fair manner.
    The results indicate that alternative research directions might be better suited to enable higher quality depth perception.
    Our data, models and code are publicly available at \url{https://vcl3d.github.io/SphericalViewSynthesis/}.
\end{abstract}

\section{Introduction}
\begin{figure}[!htbp]
    \centering
    \includegraphics[width = \linewidth]{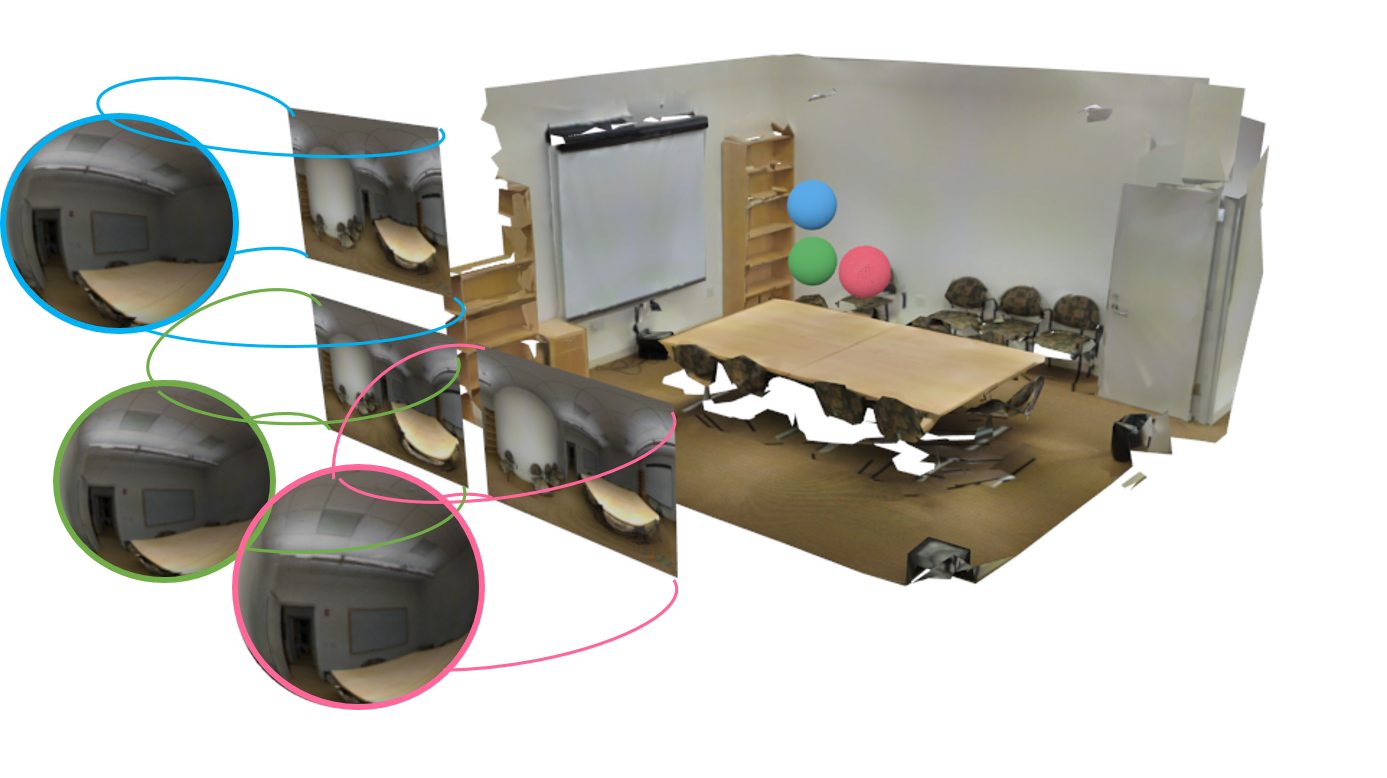}
    \caption{
    Spherical view synthesis for self-supervised depth estimation using \360 stereo images. 
    Considering spherical viewpoints within a 3D scene, we render color images from consistent baselines. 
    Starting from a central viewpoint (\textcolor{green}{green}), we explore both vertical (\textcolor{cyan}{cyan}) and horizontal (\textcolor{pink}{pink}) setups, as well as the trinocular case. 
    Indicative equirectangular projection images as observed by the 3 spherical viewpoints are presented on the left, while the 3D scene and viewpoint positions within it, on the right.}
    \label{fig:intro_trinocular_3d_scene}
\end{figure}

Data-driven approaches are producing impressive results in a variety of vision related tasks.
Convolutional neural networks (CNNs) are trained to match -- and even surpass -- human perception, managing to infer three-dimensional (3D) information solely from monocular images.
However, their performance is closely related to the availability of high quality training samples, which for certain tasks is tedious, expensive or even outright impossible.
While landmark annotations can be crowd-sourced, densely annotating images with ground truth depth values fits the latter category.
As a result, fully supervised depth learning has only been demonstrated in small scale datasets \cite{saxena2008make3d}, usually without pixel perfect depth measurements \cite{silberman2012indoor}, or otherwise in generated synthetic datasets \cite{zhang2017physically} that, nonetheless, need to overcome the synthetic-to-real domain gap. 

Naturally, a great body of work has identified this challenge and focused on overcoming it with self-supervision, using an indirect objective to infer depth, namely, view synthesis.
Accurately explaining imaged content from a different viewpoint relies on 3D information, and by extension, accurate depth.
Even though view synthesis supervision relies on a set of assumptions (diffuse materials, absence of occlusions, static scenes) that do not necessarily hold for real world acquired data, convincing depth estimation results have been presented without using any ground truth.
Earlier efforts relied on synchronized stereo cameras capturing static scenes \cite{garg2016unsupervised, godard2017unsupervised}, introducing view synthesis via inverse image warping and paving the way follow-up works.

Circumventing the need for stereo data acquisition, more recent works \cite{zhou2017unsupervised} only rely on video input for learning to infer depth.
To achieve this they learn to estimate the camera's motion jointly with estimating the observed scene's depth, and are supervised by synthesizing future and/or past views.
While an abundance of data are readily available for these structure-from-motion learning methods, they still need to overcome the violation of the static scene assumption. 

The absolute majority of this line of research has focused on traditional pinhole cameras, disregarding self-supervised depth estimation for omnidirectional input, apart from \cite{DBLP:journals/corr/abs-1811-05304}, which, however, uses a cube map (\textit{i.e.}~pinhole) representation.
Spherical view synthesis is relatively unexplored as most works focused on catadioptric or cylindrical cameras.
It is also challenging due to the inherent distortions when applied to two-dimensional images, which in turn manifest into severe self-occlusions.
Further, due to the content's spherical nature, the irregular disparity patterns that it exhibits hinder efficient learning, especially for horizontal baselines, while the singularities at the epipoles prevent coherent gradient flows when using inverse image warping. 

In this work, we explore spherical view synthesis and demonstrate its applicability for self-supervised spherical depth estimation.
Summarizing, our contributions are:
\begin{itemize}
    \item The full spherical disparity model is presented using a purely geometric derivation.
    \item A robust supervision scheme is developed for spherical view synthesis using depth-image-based rendering (DIBR) and spherical attention.
    \item Unlike inefficient and resource consuming spherical learning approaches, our network design incorporates a straightforward way to make our model aware of its spherical nature.
    \item Besides offering a large \360 stereo dataset, our work is uniquely posed to compare the effectiveness of view synthesis and direct supervision. We perform a fair and consistent evaluation and present its results.
\end{itemize}
\section{Related Work}
\textbf{Learning with spherical content:}
Applying CNNs to spherical content is accomplished by warping it to a regular grid.
\href{https://mpeg.chiariglione.org/standards/mpeg-i/omnidirectional-media-format}{MPEG-OMAF} \cite{skupin2017standardization} defines two projection formats for \360 images, the cubemap and equirectangular (ERP) projections.

While cubemaps can be straightforwardly fed into a CNN, and then re-merged back into \360 as in \cite{monroy2018salnet360}, they still suffer from cubemap distortion and discontinuity artifacts.
For the latter, cube padding \cite{cheng2018cube} can explicitly aid the network into connecting the cube faces, enabling global reasoning.
Similarly, circular padding \cite{wang2018omnidirectional} has been used when applying convolutions directly to the ERP image.

A novel direction is to bypass learning on spherical data and instead, adapt models trained on perspective images to the \360 domain.
Initially, \cite{su2017learning} regressed per row rectangular filters from the pre-trained ones, at the expense of increasing the model's size and complexity (multiple filters for a single activation map) and suffering from regression approximation.
It was recently extended \cite{su2018kernel} to transfer 2D CNN models by producing functions that map weights to each row, while preserving inter-channel information exchange, and overcoming some of the previous disadvantages, albeit still taking a model size hit (even though significantly reduced).
Another approach is adapting the input data to the \360 domain \cite{payen2018eliminating}, yet it was not demonstrated for full spherical images, but rather only for panoramic ones. 

Another direction is training rotation equivariant CNNs either using graph-based learning \cite{khasanova2017graph} or employing spectral learning approaches, with two notable works using spherical harmonics \cite{esteves2018learning} and spherical cross correlation with Fast Fourier Transforms (FFTs) \cite{cohen2018spherical} to achieve expressive training on the sphere. 
Still, their high memory footprint hinders applicability due to limited input resolutions.

As a result, more efficient approaches resorted to kernel distortion \cite{tateno2018distortion}, tangent plane kernels \cite{coors2018spherenet}, kernel resampling \cite{zhang2018saliency} or ERP specific dilations \cite{fernandez2019corners}.
However, as presented in \cite{su2018kernel}, all these approaches are valid only for the first layers, as the CNN's non-linearity distorts the pure spherical representation as the network deepens, breaking the assumptions they are designed for (i.e. the features' spherical smoothness). 
In addition, inefficient implementations \cite{zhang2018saliency} introduce problems during training (very small batch size and low run-time performance).
Instead, we resort to a more explicit and efficient solution to make the network aware of the data spherical nature, by exploiting recent research related to CNNs' capacity to self-localize their features, and also utilize spherical attention to allow for distortion aware supervision in the ERP domain.    

\textbf{Monocular self-supervised (spherical) depth:}
The seminal works of \cite{garg2016unsupervised} and \cite{godard2017unsupervised} first demonstrated that view synthesis can serve as the supervisory signal for monocular depth estimation.
This has attracted a lot of attention from the research community given the difficulty in obtaining high quality real world depth measurements.
Both \cite{garg2016unsupervised} and \cite{godard2017unsupervised} used perspective horizontal stereo data and employed either approximately \cite{garg2016unsupervised} or locally \cite{godard2017unsupervised} differentiable image warping \cite{jaderberg2015spatial} to synthesize the reconstructed views.

A novel solution was introduced by \cite{zhou2017unsupervised} that extended view synthesis supervision to unstructured video datasets by simultaneously predicting inter-frame pose.
However, learning to estimate depth purely from video breaks the static scene assumption and necessitates the use of an attention mechanism for foreground motion between consecutive frames. 
More recent iterations of this direction added scale normalization and removal of the separate pose estimation branch \cite{wang2018learning}, 3D geometric constraints between the predicted depths \cite{mahjourian2018unsupervised}, epipolar constraints \cite{prasad2019sfmlearner++}, additional feature reconstruction supervision \cite{zhan2018unsupervised}, stereo matching constraints \cite{Luo2018SVS} or explicitly used two consecutive frames as input \cite{10.1007/978-3-030-11015-4_27}.

Prevalent for all the above methods is the reconstruction loss of synthesized views via inverse warping through a stereo disparity model or explicit 3D transformations and projections.
Disregarding the challenging Lambertian surfaces assumption, inverse warping does not gracefully handle occlusions, which are only implicitly addressed (e.g. explainability/visibility masks, left-right consistencies).
This has a detrimental effect for spherical images as occlusions are magnified due to distortion.
Instead, we rely on a soft rendering approach to synthesize the supervisory views.

While a large body of work exists for traditional perspective images, scarce research has addressed depth estimation from spherical panoramas.
The most apparent issue is the unavailability of data, and thus, two concurrent works addressed \360 depth estimation by generating data via rendering existing 3D datasets.
Two baseline models were presented in \cite{zioulis2018omnidepth} after creating a large dataset of color and depth pairs using a mix of synthetic and real scenes.
Further, \cite{DBLP:journals/corr/abs-1811-05304} utilized the more recent advances in depth estimation from videos and rendered videos from a purely synthetic 3D dataset.
Still, \cite{zioulis2018omnidepth} simply applied a CNN on ERP images while \cite{DBLP:journals/corr/abs-1811-05304} explicitly used a cubemap representation and relied on previous works on perspective depth video learning, but with cubemap constraints.

Indirect supervision through spherical view synthesis has not been explored yet for learning monocular \360 depth estimation.
Previous works mainly focused on estimating depth from fisheye \cite{li2008binocular} or cylindrical \cite{zhu2001omnidirectional} stereo setups and utilized the corresponding disparity models.
For the full spherical setting a complete disparity model has not been considered as prior work only focused in extracting depth measurements and not synthesizing views.
Consequently, \360 vertical stereo setups \cite{kim20133d} were preferred due to their simpler disparity model that requires no rectification. 
Works using \360 horizontal stereo \cite{lin2018real, khasanova2017graph} relied only on horizontal disparity modeling which is sufficient to triangulate depth values after rectification.
On the other hand, horizontal spherical view synthesis introduces distortions which manifest as vertical disparity.
In this work we present and explore the complete spherical disparity model for both stereo placements under a view synthesis, self-supervised \360 depth estimation learning context. 
\section{Self-supervised Spherical Depth}
\subsection{Spherical Disparity Model}
\label{sec:spherical_disparity}
\begin{figure}[!htbp]
    \centering
    \includegraphics[width = \linewidth]{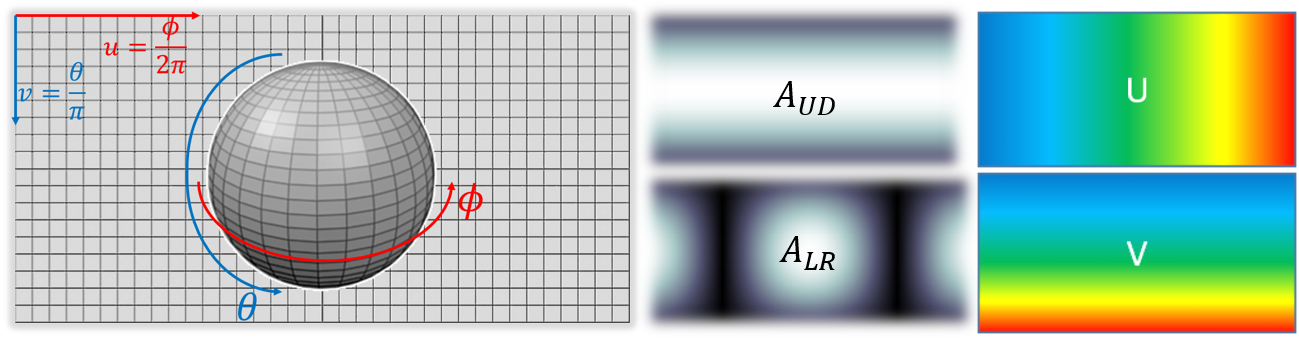}
    \caption{Spherical $\boldsymbol{\varrho} = (\phi, \theta)$ longitudinal and latitudinal coordinates aligned with the image grid's equirectangular coordinates $\mathbf{p} = (u, v)$ respectively (\textbf{left}).
    The spherical attention masks $A_{UD}$ for vertical and $A_{LR}$ for horizontal stereo placements as defined in Eq.~\ref{eq:spherical_weights} respectively (\textbf{middle}).
    Both attenuate towards the singularities, while the $A_{LR}$ also includes distortion related attenuation.
    Image grid coordinate feature maps for the horizontal ($u$) and vertical ($v$) image grid directions (\textbf{right}).}
    \label{fig:coordinates}
\end{figure}

We define a spherical image through its ERP on a 2D grid as shown in Fig.~\ref{fig:coordinates}. 
Each image's local 3D coordinate system in spherical $\boldsymbol{\rho}=(r, \phi, \theta)$ and Cartesian $\mathbf{v}=(x, y, z)$ coordinate systems are given in Eq.~\ref{eq:spherical_cartesian_coords}.
An ERP image's width $w$ and height $h$ span $w \times h := 2\pi \times \pi$ radians at the $[0,2\pi]$ and $[0,\pi]$ ranges respectively, covering a complete spherical view with $\varphi = 2\pi / w$ the horizontal and $\vartheta = \pi / h$ the vertical angular resolutions respectively. 
Columns correspond to constant longitude/azimuth ($\phi$) angles, while rows to constant latitude/elevation ($\theta$) angles.
Each pixel $\mathbf{p} = (u,v)$ can be mapped to angular spherical coordinates $\boldsymbol{\varrho} = (\phi,\theta)$ as $(u\varphi, v\vartheta)$ and vise versa.
This linear mapping between image domain pixels $\mathbf{p}$ and spherical domain angular coordinates $\boldsymbol{\varrho}$ allows for straightforward transitions between image and spherical based operations. 
We will therefore omit any explicit conversions between them in the following text.  
Contrary to perspective images, \360 depth is defined as the 3D Euclidean distance to a point, which corresponds to the radius $r$ in spherical coordinates.

\begin{gather}
\label{eq:spherical_cartesian_coords}
    \left[\begin{matrix}
        r \\
        \phi \\
        \theta
    \end{matrix}\right]
    \!\!=\!\! \left[\begin{matrix}
        (x^2 + y^2 + z^2)^{1/2} \\
        \arctan(x / z)    \\
        \arccos(y / r)
    \end{matrix}\right]\!\!,\!\!
    \left[\begin{matrix}
        x \\
        y \\
        z
    \end{matrix}\right]
    \!\!=\!\! \left[\begin{matrix}
        r\sin(\phi)\sin(\theta) \\
        r\cos(\theta)               \\
        r\cos(\phi)\sin(\theta)
    \end{matrix}\right]
\end{gather}

\begin{figure*}[!htbp]
    \centering
    \includegraphics[width=\textwidth]{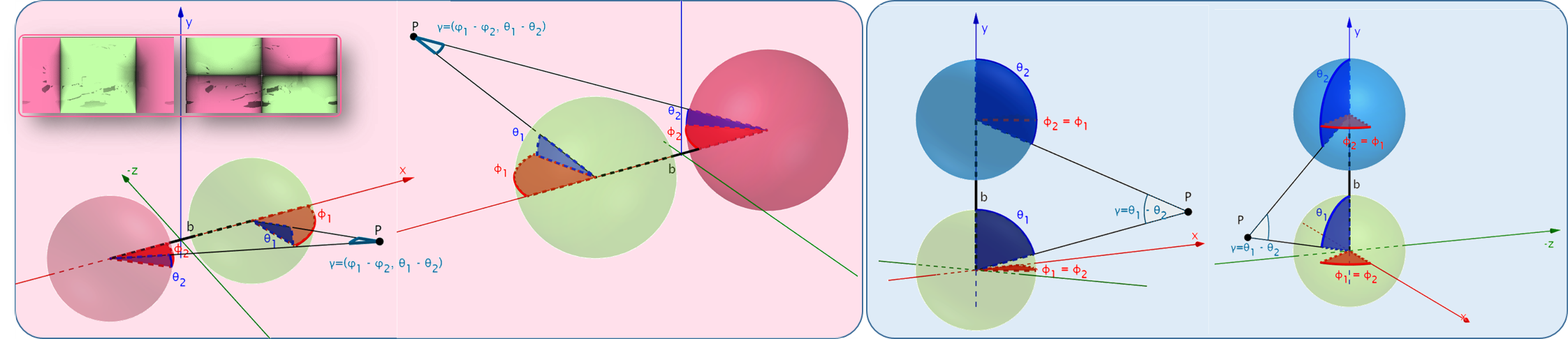}
    \caption{The spherical disparity $\gamma$ model for a horizontal (\textcolor{pink}{pink} region) and vertical (\textcolor{cyan}{cyan} region) baselines.
    Besides longitudinal disparity, the horizontal placement introduces latitudinal disparity as well, with both being a function of the estimated depth according to Eq.~\ref{eq:spherical_derivs}.
    For the vertical placement scenario, a simpler model that only includes latitudinal disparity simplifies spherical view synthesis and depth estimation.
    The top left inset illustrates the irregular sign patterns of the disparities in the horizontal stereo placement setting (negative -- \textcolor{lime}{green} and positive -- \textcolor{pink}{pink}). The left image corresponds to longitudinal ($\phi$), and the right to latitudinal ($\theta$) disparity. }
    \label{fig:spherical_geometry}
\end{figure*}

Spherical stereo considers physical displaced spherical viewpoints that image the same scene.
They are positioned with a known baseline in either horizontal or vertical placements.
Fig.~\ref{fig:spherical_geometry} shows both of these placements as well as the projection of a 3D point on each displaced viewpoint.
Disparities $\boldsymbol{\gamma} = (\gamma_{\phi}, \gamma_{\theta})$ correspond to angular differences in the angular spherical coordinates $(\phi, \theta)$ measured in radians.
Spherical disparities $\boldsymbol{\gamma}$ can be analytically derived from a source viewpoint $\mathbf{v}_{src}$ with respect to an unrotated target viewpoint $\mathbf{v}_{tgt}$ according to their baseline $\mathbf{b} = \mathbf{v}_{src} - \mathbf{v}_{tgt}$ by calculating the partial derivatives of the spherical coordinates with respect to the Cartesian ones: 

\begin{equation}
    \begin{aligned}
    \label{eq:spherical_derivs}
                \left[\begin{matrix}
                    \partial{r}     \\
                    \partial{\phi}  \\
                    \partial{\theta}\\
                \end{matrix}\right]
                \!\!\!=\!\!\!\left[\begin{matrix}
                    \sin(\phi)\sin(\theta) & \cos(\theta) & \cos(\phi)\sin(\theta) \\
                    \textcolor{red}{\frac{\cos(\phi)}{r\sin(\theta)}} & \textcolor{blue}{0} & \frac{-\sin(\phi)}{r\sin(\theta)}                   \\ \textcolor{red}{\frac{\sin(\phi)\cos(\theta)}{r}} & \textcolor{blue}{\frac{-\sin(\theta)}{r}} & \frac{\cos(\phi)\cos(\theta)}{r}
                \end{matrix}\right]
                \!\!\!
                \left[\begin{matrix}
                    \partial{x} \\
                    \partial{y} \\
                    \partial{z}
                \end{matrix}\right]\!\!\!
    \end{aligned}
\end{equation}
These link Cartesian displacements, \textit{i.e.}~the baseline $\mathbf{b} = (\mathbf{d}x, \mathbf{d}y, \mathbf{d}z)$, to angular displacements on the sphere, \textit{i.e.}~disparity $\boldsymbol{\gamma} = (\mathbf{d}\phi, \mathbf{d}\theta)$, through the radius, \textit{i.e.}~depth, $r$.

For horizontal stereo along the $\textcolor{red}{x}$ \textcolor{red}{axis} with a baseline $\mathbf{b}_x\!\!=\!\!(\mathbf{d}x, 0, 0)$ it is $\textcolor{red}{\boldsymbol{\gamma_{horiz}} = \mathbf{d}_x (\partial{\phi} / \partial{x}, \partial{\theta} / \partial{x})}$ while for vertical stereo with a baseline $\mathbf{b}_y\!\!=\!\!(0, \mathbf{d}y, 0)$ along the $\textcolor{blue}{y}$ \textcolor{blue}{axis} it is $\textcolor{blue}{\boldsymbol{\gamma_{vert}} = \mathbf{d}_y (\partial{\phi} / \partial{y}, \partial{\theta} / \partial{y})}$.
Evidently, the disparity model for vertical placements is simpler, as there is no longitudinal disparity and thus, the pixels reproject to the same vertical scan path.
However, for horizontal placements the reprojected pixels lie on epipolar curves on the ERP domain which are sinusoidal, resulting in a more complex disparity model with displacements along both angular directions.

\subsection{Depth-image-based rendering}
As presented in Sec.~\ref{sec:spherical_disparity}, the angular disparity $\boldsymbol{\gamma}$ for an ERP pixel $\boldsymbol{\varrho} = (\phi, \theta)$ is a function of its depth $r$ and the baseline $\mathbf{b}$ between the viewpoints.
Consequently, we can transform pixel coordinates from a source ERP image $I_s$ to a target ERP image $I_t$ given the source's depth map $D_s$ using an angular pixel displacement function $\Gamma$:
\begin{equation}
    \boldsymbol{\varrho}_{t}=\Gamma_{s\rightarrow t}(D_{s}, \boldsymbol{\varrho}_{s}, \mathbf{b}_{s\rightarrow t}) = \boldsymbol{\varrho}_{s} - \gamma(D_{s}, \boldsymbol{\varrho}_{s}, \mathbf{b}_{s\rightarrow t}).
\end{equation}
It should be noted that for horizontal stereo, the longitudinal disparity wraps around the sphere. This corresponds to a modulo operation, which is omitted to simplify notation.

Under a traditional inverse warping approach, the target image would be bilinearly sampled to synthesize the source view and supervise learning through the reconstructed source view.
Yet, this approach cannot easily handle occluded regions or non-linear mappings which are prevalent in the sphere.
Indeed, the ERP distortions are responsible for many-to-one as well as one-to-many pixel mappings, a fact that is more pronounced in wider baselines that are a necessity for higher accuracy in farther depths.
Furthermore, wider baselines produce noticeable occlusions, especially for spherically imaged content.

\begin{figure}[!htbp]
    \centering
    \includegraphics[width = \linewidth]{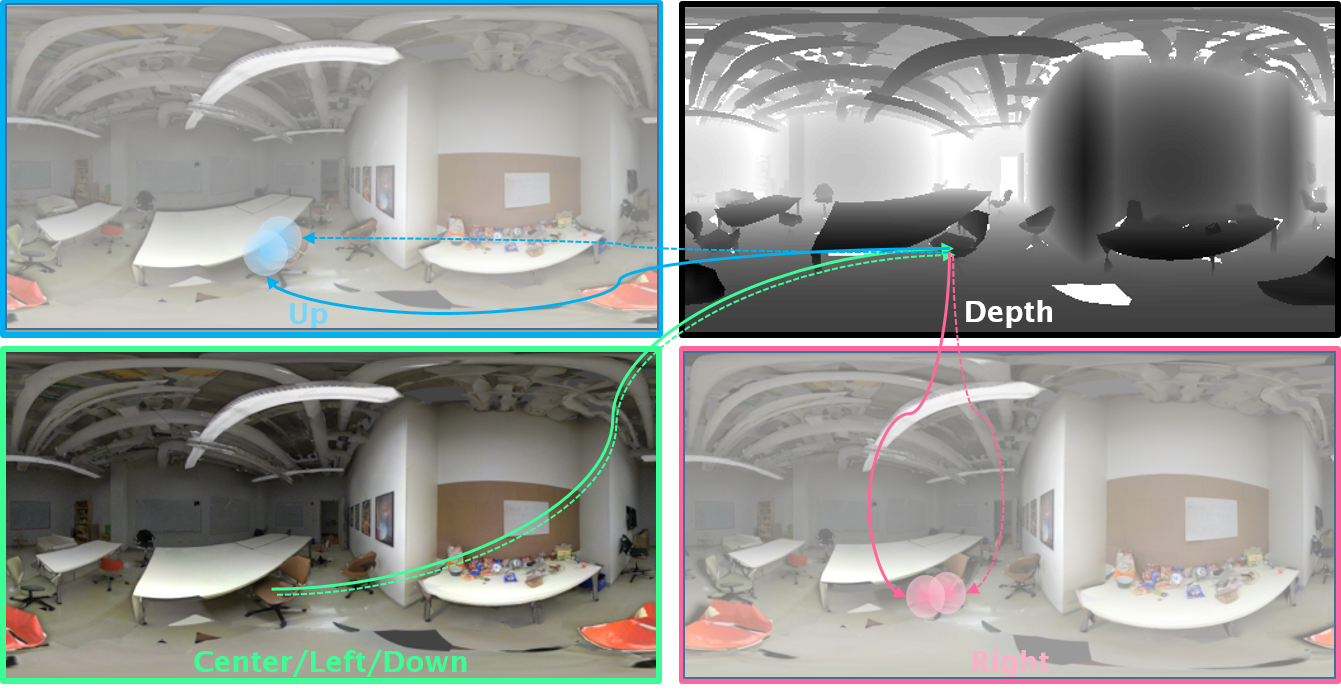}
    \caption{
    Depth-image-based rendering view synthesis. 
    For each pixel, its reprojection to another viewpoint (in \textcolor{pink}{horizontal} or \textcolor{cyan}{vertical} stereo placement) is calculated via the estimated depth map. 
    For each pixel a weighted splat is added to an empty canvas around its immediate reprojection neighborhood. 
    Subsequent to their accumulation, a  normalization step produces the final rendering.
    In this way, occlusions and irregular pixels mappings are handled gracefully, allowing for the use of view synthesis as a supervision objective, even for severe distortion areas. }
    \label{fig:splatting}
\end{figure}

In order to enable learning through view synthesis for spherical stereo we use a soft locally differentiable rendering approach (DIBR) that involves splatting the contributions of each source image pixel to an empty target canvas $\hat{\textbf{I}}_t$ (Fig.~\ref{fig:splatting}).
The splatted coordinates are derived by $\Gamma$ and are a function of the source depth map $D_s$.
Local differentiability is ensured by neighborhood based bilinear splatting, while soft rendering relies on weighted contribution accumulation in the target image \cite{tulsiani2018layer}.

In more detail, each source pixel $\boldsymbol{\varrho}_s$ contributes to four target pixels $\boldsymbol{\varrho}_t^{\mathcal{N}} : \{\boldsymbol{\varrho}_t^{tl}, \boldsymbol{\varrho}_t^{tr}, \boldsymbol{\varrho}_t^{bl}, \boldsymbol{\varrho}_t^{br}\}$ comprising a neighborhood $\mathcal{N}$ created through floor and ceiling operations on the target pixel's $\boldsymbol{\varrho}_t$ coordinates.
A bilinear weight $\beta(\boldsymbol{\varrho}_t^{\mathcal{N}}, \boldsymbol{\varrho}_t)$ is associated with each of them.
The contributions of all source pixels are accumulated on the target image via scattering operations and additionally weighted by a depth attenuation factor $\alpha(\boldsymbol{\varrho}_s, D) = e^{-D(\boldsymbol{\varrho}_s) / d_{max}}$, with $d_{max}$ a pre-selected maximum depth value.
Each source pixel's contribution to the target image canvas $\hat{\textbf{I}}_t$ is weighted by $w(\boldsymbol{\varrho}_s) = \alpha(\boldsymbol{\varrho}_s, D)\beta(\boldsymbol{\varrho}_t^{\mathcal{N}}, \boldsymbol{\varrho}_t)$. 
Additionally, the weights themselves are also splatted in a target weight canvas $\hat{W}_t$.

In this way, soft z-buffering is enforced and the target view $\tilde{\mathbf{I}}_t$ is synthesized, after a normalization operation that divides the splatted color canvas with the splatted weight canvas in an element-wise fashion: $\tilde{\mathbf{I}}_t = \hat{\mathbf{I}}_t \oslash (\hat{W}_t + \epsilon)$, $\epsilon$ being a small numerical stability constant.
This allows for backpropagation to the occluded areas whose view synthesis contributions and gradients are weighted according to a viewpoint proximity criterion. 
Besides gracefully handling occlusions, this splatting based view synthesis can accommodate many-to-one pixel mappings.
While one-to-many pixel mappings are not supported, they do not need to be explicitly handled as the canvas will be empty in those regions where no source pixel contribution landed.
This way, a binary mask $M_t = \hat{W}_t < \varepsilon$ can be calculated that masks empty canvas areas.
On the contrary, when using inverse warping, either ground truth depth for z-testing, or an attention mechanism is required to prevent false supervision and destabilizing gradient backpropagation.

\subsection{CoordNet}
\label{sec:coordnet}
\begin{figure*}[!htbp]
    \centering
    \includegraphics[width = \textwidth]{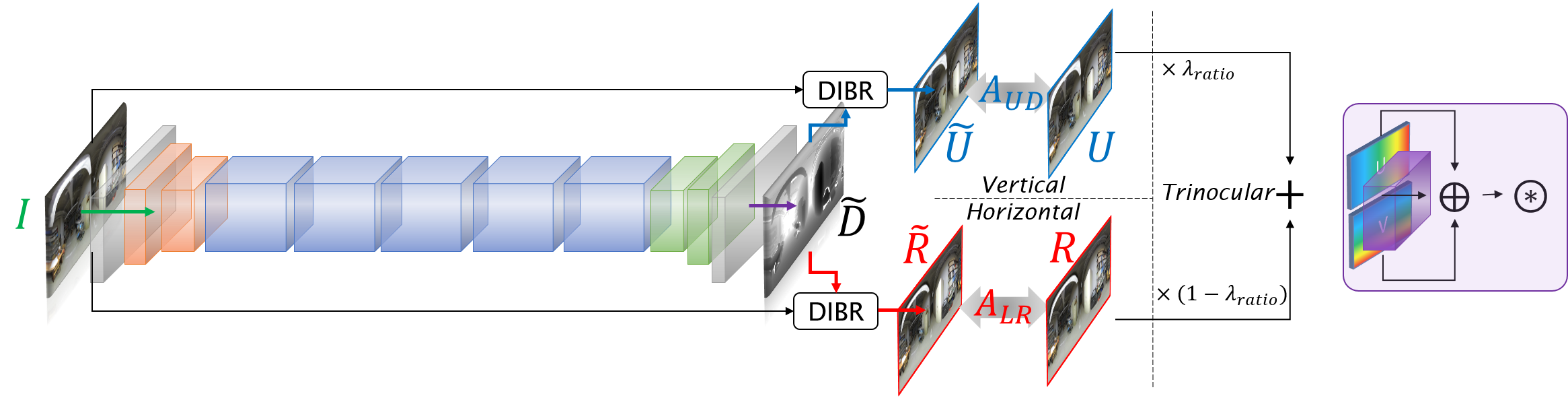}
    \caption{
    CoordNet and its view synthesis based supervision scheme. 
    An ERP depth map $\tilde{D}$ is predicted using a single monocular ERP color image $\tilde{\mathbf{I}}$. 
    Through the estimated depth, we synthesize stereo viewpoints in vertical (up -- $\tilde{\mathbf{U}}$) or horizontal (right -- $\tilde{\mathbf{R}}$) baselines.
    These are supervised using a photometric consistency error using the original viewpoints $\mathbf{U}$ and $\mathbf{R}$ for the up and right reconstructions, as well as placement specific attention maps, $A_{UD}$ and $A_{LR}$ respectively.
    The trinocular self-supervised scenario is also considered using a blending factor $\lambda_{ratio}$ to balance loss between the two different viewpoint reconstructions.
    CoordNet utilizes CoordConvs for all its convolutional layers as shown on the right.
    Each incoming feature map is concatenated with the horizontal $u$ and vertical $v$ coordinate maps of its resolution before fed into the convolution operation.}
    \label{fig:network}
\end{figure*}

\textbf{Architecture:}
Our network, CoordNet, illustrated in Fig.~\ref{fig:network}, is designed to be efficient in learning with spherical data, minimizing memory consumption and maximizing inference speed compared to other approaches for \360 learning.
Our lightweight backbone architecture is inspired by \cite{johnson2016perceptual} but we replace traditional residual blocks with pre-activated ones \cite{he2016identity} and utilize ELU \cite{shah2016deep} activations instead of RELU \cite{nair2010rectified} and batch normalization \cite{ioffe2015batch}.

We introduce \360 awareness implicitly within our model by utilizing the recently introduced coordinate convolutions \cite{liu2018intriguing}.
Each input feature map is concatenated with two additional feature maps that represent its grid coordinates in the two dimensional grid.
These extra features allow the network to learn the spatial context, which in our case is the ERP domain.
CoordNet has minimal memory overhead compared to spectral or model transference approaches, which only scales with feature resolution and the number of convolutional layers.
Additionally, in terms of run-time performance, the processing overhead is lower than kernel based approaches that involve trigonometry calculations for warping features or weights. 

Unlike other stereo self-supervised learning approaches, we resort to predicting depth directly instead of disparities, and use Eq.~\ref{eq:spherical_derivs} to calculate them for view synthesis.
This allows for a more general spherical view synthesis model that can facilitate both vertical and horizontal stereo placements.
For the vertical case, a direct disparity estimation is equivalent to depth estimation, but for the horizontal one, it touches on an important weakness of CNNs: their inability to simultaneously regress spatially varying positive and negative values.
Longitudinal disparities for horizontal stereo are of opposing signs at the front and back looking directions.
Moreover, latitudinal disparities, in the same placement, follow spatially varying sign patterns, depicted in Fig.~\ref{fig:spherical_geometry}, further magnifying the problem.
While a solution would be to predict absolute values and explicitly enforce correct signs this was not the case in our experiments as training did not manage to converge.
Since the longitudinal and latitudinal disparities are correlated, directly predicting the first would make the estimation of the second possible, but only after transitioning to depth, yet this only strengthens the choice of regressing depth directly.

\textbf{Supervision:}
CoordNet is self-supervised by a depth driven photometric image reconstruction loss as well as a depth smoothness prior:
\begin{equation}
\label{eq:loss}
    \mathcal{L}_{total} = \lambda_{recon}  \mathcal{L}_{recon} + \lambda_{smooth} \mathcal{L}_{smooth},
\end{equation}
where $\lambda_{recon}$ and $\lambda_{smooth}$ are weights that sum up to one. 
Our reconstruction loss uses a standard photometric loss as presented in \cite{godard2017unsupervised}, which is also used in most self-supervised monocular depth estimation methods:
\begin{equation*}
\mathcal{L}_{photo}(\mathbf{p}) \! = \! \eta \mathcal{L}_{D}(\mathbf{I}_t^M \! (\mathbf{p}), \tilde{\mathbf{I}}_t^M \! (\mathbf{p})) + (1-\eta)\big\vert\mathbf{I}_t^M \! (\mathbf{p})- \tilde{\mathbf{I}}^M_t \! (\mathbf{p})\big\vert.
\end{equation*}
It combines the L1 penalty function with structural dissimilarity $\mathcal{L}_{D}$, under a relative weighting factor $\eta$. 
The superscript $M$ denotes multiplication with the binary mask $M_t$.

While previous ERP domain learning approaches \cite{zioulis2018omnidepth} used uniform supervision on the ERP image, such an approach will greatly bias higher quality predictions towards the more distorted areas.
Instead, we explicitly use a spherically weighted attention mechanism to uniformly aggregate errors and gradients on the sphere, instead of on the distorted ERP image.
We use an attention weight matrix $A$ defined on the ERP domain in two different variants:
\begin{equation}
\label{eq:spherical_weights}
    A(\boldsymbol{\varrho}) = \begin{cases}
    \vert\sin(\theta)\vert, \quad \quad \quad \, \, \, \text{for vertical stereo,} \\
    \vert\sin(\phi)\vert\vert\sin(\theta)\vert, \, \text{for horizontal stereo}.
    \end{cases}
\end{equation}
These weight maps, as illustrated in Fig.~\ref{fig:coordinates}, eliminate the effect of the epipole singularities as the contributions of the areas around the singularities tend to zero.
For the vertical case, they coincide with the distortion attenuation factor $\sin(\theta)$ but for the horizontal case the corresponding singularity attenuation term $\sin(\phi)$ is added.
Hence, the total reconstruction loss is the spherically weighted mean photometric error of all valid pixels:
\begin{equation}
    \mathcal{L}_{recon} = \frac{1}{\sum_{\mathbf{p}}M_t(\mathbf{p})} \sum_{\mathbf{p}} A(\mathbf{p})M_t(\mathbf{p})\mathcal{L}_{photo}(\mathbf{p}).
\end{equation}
We also impose a smoothly varying prior on the predicted signal.
However, defining smoothness on the sphere is challenging and naive approaches like applying finite element gradient operators \cite{zioulis2018omnidepth} in the ERP domain will not succeed in enforcing smoothness correctly, as spherical depth inherently varies spatially even for flat surfaces.
As an alternative, we enforce a smoothness constraint on the deprojected Cartesian coordinates $\mathbf{v} \! = \! (x, y, z)$ for each predicted pixel $(r \! = \! \tilde{D}(\mathbf{p}), \phi, \theta)$, by minimizing the following weighted total variation term, using central differences:
\begin{equation}
    \mathcal{L}_{smooth}\!\! =\!\!\bar{A}(\mathbf{p}) e^{\vert\!\vert\nabla\mathbf{I}_s(\mathbf{p})\vert\!\vert_2} \sqrt{(\nabla_u \mathbf{v}(\mathbf{p}))^2\!\!+ \!\!(\nabla_v \mathbf{v}(\mathbf{p}))^2}.
\end{equation}
The weighting term $\bar{A}(\mathbf{p}) = \mathbf{1} - A(\mathbf{p})$ more heavily enforces smoothness on the distorted regions. 
A color guidance weighted factor is also used in order to establish correlated depth and color gradients.
Thus, smoothness on the ERP domain is ensured via the Eq.~\ref{eq:spherical_cartesian_coords} deprojection functions.

\section{Results}
\begin{figure*}[!htbp]
    \centering
    \includegraphics[width = \textwidth]{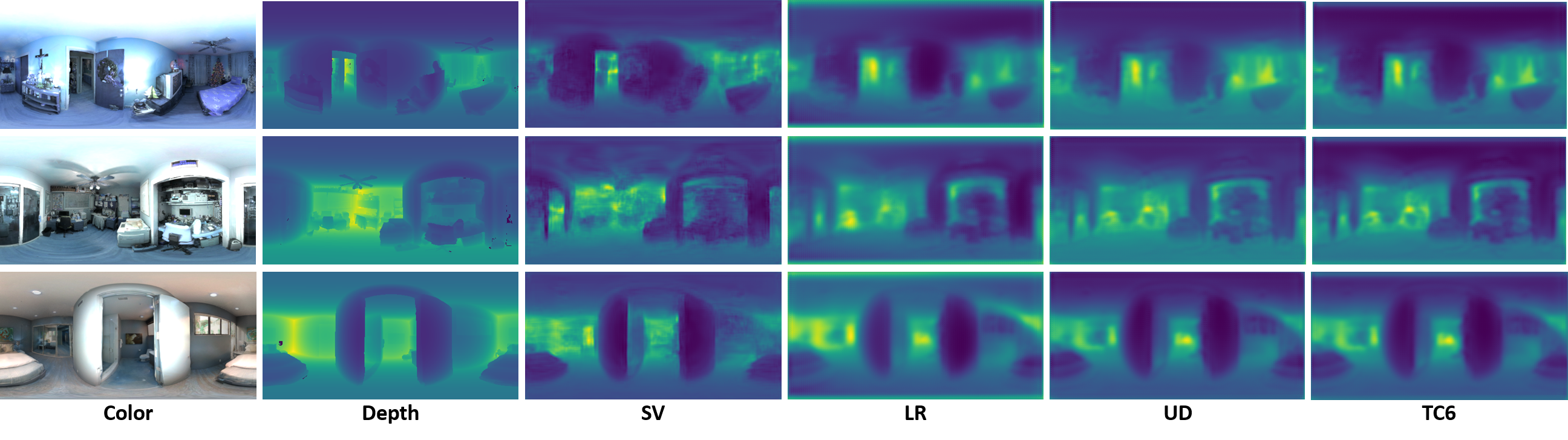}
    \caption{
    Qualitative results of each category of trained models (TC6 was chosen as it is the best performing trinocular model).
    From left to right: the input color image, the ground truth depth from \cite{zioulis2018omnidepth}, the fully supervised prediction (SV), the self-supervised predictions of horizontal (LR), vertical (UD) and trinocular (TC6) placements.
    Additional examples can be found in our supplementary material. }
    \label{fig:qualitative}
\end{figure*}

\textbf{Dataset:}
Given the unavailability of stereo \360 datasets, we take a similar approach to \cite{zioulis2018omnidepth} and render panoramas from displaced viewpoints in both vertical and horizontal placements as shown in Fig.~\ref{fig:spherical_geometry}.
We use Blender\footnote{Blender uses different longitudinal and latitudinal ranges ($
[\frac{-3\pi}{2}, \frac{\pi}{2}]$ and $[\frac{-\pi}{2}, \frac{\pi}{2}]$ respectively), therefore Eq.~\ref{eq:spherical_cartesian_coords}, Eq.~\ref{eq:spherical_derivs} and Eq.~\ref{eq:spherical_weights} get modified accordingly using trigonometric reflections.} and set the baseline for both placements to $0.26$m, which is a reasonable distance to get high quality results for indoor scenes, which is the context of the rendered 3D datasets used in \cite{zioulis2018omnidepth}.
However, unlike \cite{zioulis2018omnidepth}, we use the official train, validation and test splits of Matterport3D \cite{chang2017matterport3d} and Stanford2D3D \cite{2017arXiv170201105A} (fold\#1).
In this way, our test set is sufficiently different from our train set, and at least quadruple the size of the test set used in \cite{zioulis2018omnidepth}. 
Further, SunCG \cite{song2017semantic} is only used during training and validation, but not during testing as our focus is to assess applicability in real world settings.

\textbf{Implementation Details:}
We implement our network in PyTorch \cite{paszke2017automatic}, initialize its weights using \cite{glorot2010understanding}, and train all our models for $30$ epochs using a fixed learning rate of $10^{-4}$ and a batch size of $16$. 
Across all experiments we use a fixed seed for all the involved random generators to guarantee consistency.
We use the AdaBound \cite{Luo2019AdaBound} optimizer with a convergence speed of $2 \! \times \! 10^{-3}$ and a final target SGD learning rate of $10^{-3}$.
The weights of Eq.~\ref{eq:loss} are set to $\lambda_{recon}\!\!=\!\!0.95$ and $\lambda_{smooth}\!\!=\!\!0.05$.
Inline with prior work, the photometric error is balanced by $\eta\!\!=\!\!0.85$ and a box filter with a kernel size of $5$ is used for the SSIM calculations.

\textbf{Metrics:}
We use traditional depth evaluation metrics \cite{eigen2014depth}, but with a notable difference.
While previous works on \360 depth estimation \cite{zioulis2018omnidepth, DBLP:journals/corr/abs-1811-05304} used these metrics in the ERP domain, they did not take into account its distortion.
As a result, distorted areas were given higher precedence in the error calculation.
We adapt the absolute relative error, squared relative error, RMSE and RMSLE to use weighted calculations for each pixel using the first case of Eq.~\ref{eq:spherical_weights} in order to alleviate the effect of ERP distortion in our evaluation.
However, the percentile threshold metrics require a different approach.
Instead of densely sampling the ERP, we sample the sphere using an $\mathcal{S}^2$ generalized spiral set \cite{saff1997distributing} with $N\!=\!0.25\!\times\!w\!\times\!h$ points. 
Consequently, the percentile thresholds are only calculated for these spiral points.

\textbf{Stereo placement analysis:}
First we seek to assess which stereo placement is more efficient for view synthesis based depth estimation learning.
We train two variants of the network described in Sec.~\ref{sec:coordnet}.
For the vertical variant (referred to as UD, \textit{i.e.}~up-down) we supervise using the up view (displaced on the $y$ axis) while the network is fed the down/central image.
Similarly, for the horizontal variant (LR, \textit{i.e.}~left-right), we supervise using the right view.
Table \ref{tab:all-metrics} shows that both converge at about the same epoch, and that UD achieves higher performance.
Intuitively this is attributed to the simpler disparity model.
Nevertheless, another important factor is that an UD model does not suffer from the prevalent horizontal distortions.
Due to this reason, an UD variant can be trained with inverse warping as the view synthesis method, while for the LR variant, convergence with inverse warping was not possible.

\textbf{Complementarity analysis:}
Next, we seek to understand whether these two placements are complementary.
We train another model using trinocular (referred to as TC) supervision that infers a single depth map from the central view and is jointly supervised by the reconstruction of both the up and right images, as seen in Fig.~\ref{fig:network}.
We explore the effect of blending both view synthesis supervisions by adding a ratio parameter to combine their losses $\mathcal{L}_{recon}\!\!=\!\!\lambda_{ratio}\mathcal{L}_{recon}^{UD} + (1\!-\!\lambda_{ratio})\mathcal{L}_{recon}^{LR}$.
We train $4$ variants of the TC network with a $0.2$ step for $\lambda_{ratio}$ and name them by suffixing TC with the ratio's decimal.
The results are also presented in Table \ref{tab:all-metrics} with the color coded interpolation for each metric illustrating the transition from the best to worst, as we move from LR ($\lambda_{ratio}\!\!=\!\!0)$ to UD ($\lambda_{ratio}\!\!=\!\!1)$.
Interestingly, TC4 indicates that there exist blending factors that will not allow the model to learn a good enough representation as single viewpoint supervisions do.
We further observe that performance increases as the ratio increases towards the simpler disparity model.
Nonetheless, while UD achieves best performance with respect to outlier predictions (as indicated by the RMS metrics), we find that the slower convergence of TC6 results in a more robust model, offering a compromise for overall performance, attributed to the harder to optimize for, right view reconstruction.

\begin{table}[!htbp]
\centering
\caption{Best performing snapshots (reached at the corresponding epoch on the right) of our trained models. 
Relative  performance for the self-supervised methods is color coded to showcase the gradual transition from LR to UD via the different blending factors of TC.
Lower is better for light blue metrics, while for the darker accuracies $\delta_i < 1.25^i$ higher is better.
}
\label{tab:all-metrics}
\small\addtolength{\tabcolsep}{-5pt}
\begin{tabular}{@{}ccccccccc@{}}
    & \cellcolor[HTML]{DAE8FC}\begin{tabular}[c]{@{}c@{}}Abs\\ Rel\end{tabular} & \cellcolor[HTML]{DAE8FC}\begin{tabular}[c]{@{}c@{}}Sq\\ Rel\end{tabular} & \cellcolor[HTML]{DAE8FC}RMSE  & \cellcolor[HTML]{DAE8FC}RMSLE & \cellcolor[HTML]{CBCEFB}$\delta_1$     & \cellcolor[HTML]{CBCEFB}$\delta_2$     & \cellcolor[HTML]{CBCEFB}$\delta_3$     & Epoch                      \\ \midrule
SV  & 0.138                                                                     & \textbf{0.091}                                                           & \textbf{0.473}                & \textbf{0.184}                & \textbf{82.4\%}                & \textbf{95.9\%}                & \textbf{98.5\%}                & \cellcolor[HTML]{FFC702}24 \\ \midrule
LR  & \cellcolor[HTML]{FFCCC9}0.143                                             & \cellcolor[HTML]{FFCCC9}0.129                                            & \cellcolor[HTML]{FFCCC9}0.639 & \cellcolor[HTML]{FFCCC9}0.230 & \cellcolor[HTML]{FD6864}58.1\% & \cellcolor[HTML]{FD6864}88.2\% & \cellcolor[HTML]{FFCCC9}96.5\% & \cellcolor[HTML]{FFCB2F}18 \\
TC2 & \cellcolor[HTML]{9AFF99}0.132                                             & \cellcolor[HTML]{9AFF99}0.117                                            & \cellcolor[HTML]{FFCCC9}0.606 & \cellcolor[HTML]{FFCCC9}0.216 & \cellcolor[HTML]{FFCCC9}61.3\% & \cellcolor[HTML]{FFCCC9}89.3\% & \cellcolor[HTML]{FFCCC9}96.1\% & \cellcolor[HTML]{FFCB2F}20 \\
TC4 & \cellcolor[HTML]{FD6864}0.199                                             & \cellcolor[HTML]{FD6864}0.154                                            & \cellcolor[HTML]{FD6864}0.651 & \cellcolor[HTML]{FD6864}0.250 & \cellcolor[HTML]{34FF34}65.8\% & \cellcolor[HTML]{67FD9A}91.2\% & \cellcolor[HTML]{34FF34}96.7\% & \cellcolor[HTML]{FFCB2F}17 \\
TC6 & \cellcolor[HTML]{32CB00}0.129                                             & \cellcolor[HTML]{32CB00}0.112                                            & \cellcolor[HTML]{9AFF99}0.580 & \cellcolor[HTML]{32CB00}0.209 & \cellcolor[HTML]{34FF34}65.1\% & \cellcolor[HTML]{32CB00}91.3\% & \cellcolor[HTML]{32CB00}97.0\% & \cellcolor[HTML]{CD9934}28 \\
TC8 & \cellcolor[HTML]{9AFF99}0.133                                             & \cellcolor[HTML]{9AFF99}0.117                                            & \cellcolor[HTML]{34FF34}0.578 & \cellcolor[HTML]{32CB00}0.209 & \cellcolor[HTML]{34FF34}65.4\% & \cellcolor[HTML]{9AFF99}91.0\% & \cellcolor[HTML]{34FF34}96.9\% & \cellcolor[HTML]{FFCE93}16 \\
UD  & \cellcolor[HTML]{9AFF99}0.134                                             & \cellcolor[HTML]{9AFF99}0.119                                            & \cellcolor[HTML]{32CB00}0.571 & \cellcolor[HTML]{32CB00}0.208 & \cellcolor[HTML]{32CB00}66.4\% & \cellcolor[HTML]{9AFF99}90.8\% & \cellcolor[HTML]{34FF34}96.8\% & \cellcolor[HTML]{FFCE93}16 \\ \bottomrule
\end{tabular}
\vspace{-0.15in}
\end{table}

\textbf{Self-supervision status:}
Given that we rendered/synthesized our data, we are in the unique position of being able to directly and fairly compare view synthesis self-supervision and direct supervision.
Most others self-supervised works resort to view synthesis supervision because no high quality depth ground truth data are available.
While datasets with laser scanner depth data exist, they are usually sparse, and/or of limited test samples.
On the other hand, synthetic datasets that offer high quality depth renders, do not need to render stereo viewpoints, and consequently, this comparison has not been done before.
Further, even if it is possible to perform this comparison with synthetic data, applicability to real world scenes is the ultimate goal, which our dataset supports in assessing.

We train our network modifying only the loss function and directly supervising with ground truth depth maps.
We use the BerHu loss \cite{laina2016deeper} and refer to the fully supervised train as SV.
Table \ref{tab:all-metrics} clearly shows the superiority of a fully supervised approach compared to stereo self-supervision, providing food for thought and poses interesting dilemmas.

\begin{figure}[!htbp]
    \centering
    \includegraphics[width = \linewidth]{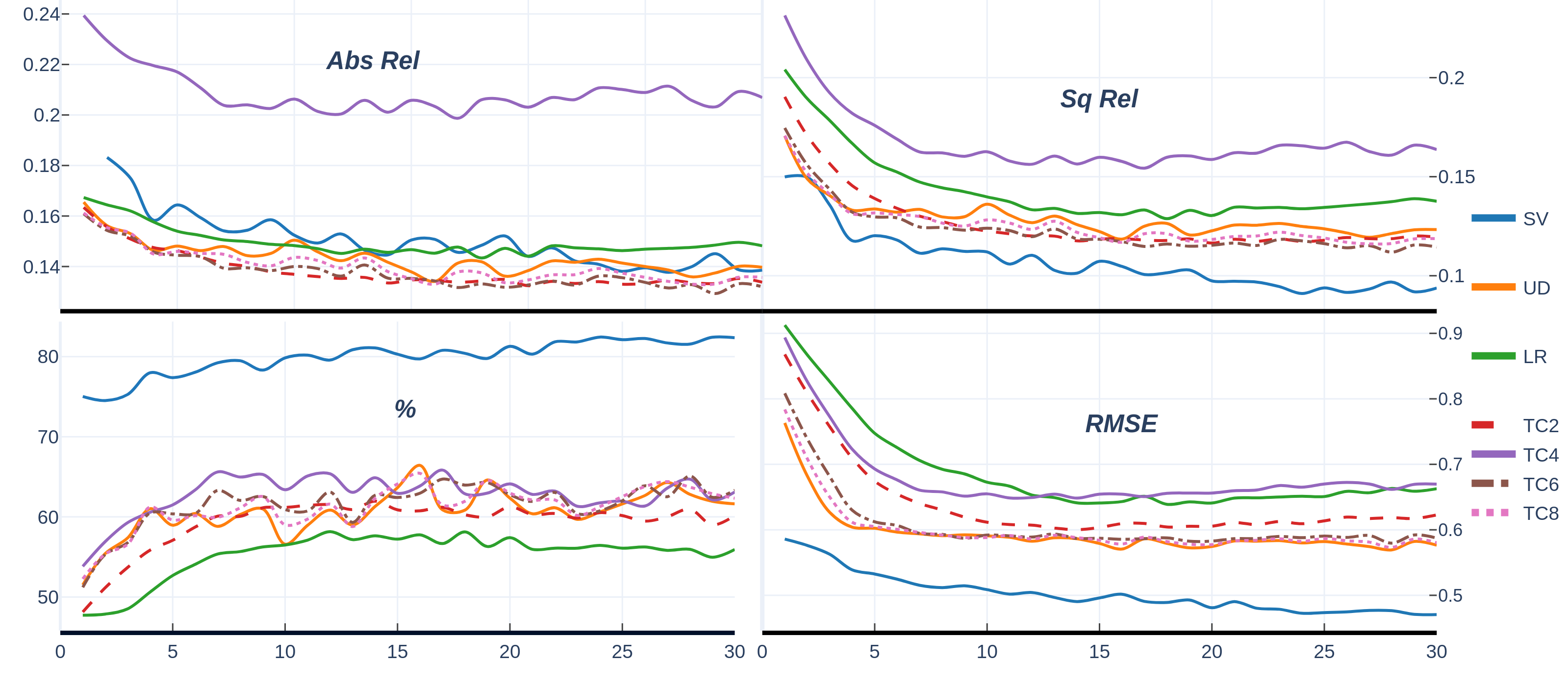}
    \caption{Test set metrics for each epoch for all the conducted experiments. Left to right, top to bottom: Absolute relative error, Squared relative error, $\delta_1 < 1.25$ accuracy, RMSE.}
    \label{fig:across_epochs}
\end{figure}

\textbf{Convergence analysis:}
We additionally offer a detailed analysis for the convergence behaviour of all variants as Table \ref{tab:all-metrics} only reported the best performing snapshots.
Fig.~\ref{fig:across_epochs} plots the results of four metrics on the whole test set across epochs.
It further signifies the importance of direct supervision as it is observed that it consistency improves its predictions.
At the same time, UD plateaus while LR is unable to converge further and instead loses performances across epochs, both after around the middle of the training duration, where they achieve their best performing state. 
The good TC variants showcase more stable training and consistently higher quality performance, contrary to UD which fluctuates more, albeit achieving a high quality minima.

\textbf{State-of-the-art comparison:}
We compare our stereo-based learning approach to a recent video-based one \cite{DBLP:journals/corr/abs-1811-05304}.
Since \cite{DBLP:journals/corr/abs-1811-05304} similarly renders a sequence dataset using SunCG scenes (PanoSunCG), we train a SunCG only variant of our TC6 model (SCG-TC6).
The first two rows of Table \ref{tab:panosuncg_metrics} compare both methods on the PanoSunCG test set.
Since \cite{DBLP:journals/corr/abs-1811-05304} does not provide a publicly available model, and only offers a quantized dataset of significantly smaller variance than ours (our test set alone uses twice as many scenes as the PanoSunCG train and test set combined), the final row of Table \ref{tab:panosuncg_metrics} presents our model's quantitative performance on our SunCG test set.
While the performance of \cite{DBLP:journals/corr/abs-1811-05304} is slightly better on PanoSunCG, our model achieves much higher quality results in our more diverse test set.

\begin{table}[!htbp]
\vspace{-0.05in}
\centering
\caption{SunCG \& PanoSunCG Comparison Results}
\label{tab:panosuncg_metrics}
\small\addtolength{\tabcolsep}{-4pt}
\begin{tabular}{@{}cccccccc@{}}
\toprule
\multicolumn{1}{l}{} & \cellcolor[HTML]{DAE8FC} \begin{tabular}[c]{@{}c@{}}Abs\\ Rel\end{tabular} & \cellcolor[HTML]{DAE8FC} \begin{tabular}[c]{@{}c@{}}Sq\\ Rel\end{tabular} & \cellcolor[HTML]{DAE8FC} RMSE & \cellcolor[HTML]{DAE8FC} RMSLE & $\cellcolor[HTML]{CBCEFB} \delta_1$ & $\cellcolor[HTML]{CBCEFB} \delta_2$ & $\cellcolor[HTML]{CBCEFB} \delta_3$ \\ \midrule
\cite{DBLP:journals/corr/abs-1811-05304} & \textbf{0.337} & \textbf{0.196} & \textbf{0.337} & 0.611 & \textbf{64.7\%} & \textbf{82.9\%} & \textbf{89.9\%} \\
SCG-TC6 & 0.371 & 0.440 & 0.843 & \textbf{0.421} & 56.2\% & 78.4\% & 87.8\% \\ \midrule
SCG-TC6 & 0.185 & 0.123 & 0.491 & 0.215 & 72.2\% & 89.5\% & 92.0\%
\\
\bottomrule
\end{tabular}
\vspace{-0.1in}
\end{table}

\textbf{CoordConv:}
Finally, we perform an ablation analysis starting with the effect of CoordConv.
We train UD and LR using standard convolutions and report the results in Table \ref{tab:coordconv_ablation}.
We observe that CoordConvs clearly boost the performance in an UD placement but it is harder to determine a similar finding for LR.
The discrepancy in RMSE and RMSLE indicate that there is a gain for closer distances (which RMSLE favors) compared to far ones (that RMSE favors), similarly indicated by the discrepancy in the relative metrics (squared against absolute).

\begin{table}[!htbp]
\centering
\caption{CoordConv Ablation Results}
\label{tab:coordconv_ablation}
\small\addtolength{\tabcolsep}{-4pt}
\begin{tabular}{@{}cccccccc@{}}
\toprule
\multicolumn{1}{l}{} & \cellcolor[HTML]{DAE8FC} \begin{tabular}[c]{@{}c@{}}Abs\\ Rel\end{tabular} & \cellcolor[HTML]{DAE8FC} \begin{tabular}[c]{@{}c@{}}Sq\\ Rel\end{tabular} & \cellcolor[HTML]{DAE8FC} RMSE & \cellcolor[HTML]{DAE8FC} RMSLE & $\cellcolor[HTML]{CBCEFB} \delta_1$ & $\cellcolor[HTML]{CBCEFB} \delta_2$ & $\cellcolor[HTML]{CBCEFB} \delta_3$ \\ \midrule
LR & 0.143 & \textbf{0.129} & \textbf{0.639} & 0.230 & 58.1\% & 88.2\% & \textbf{96.5\%} \\
w/o CC & \textbf{0.141} & 0.138 & 0.663 & \textbf{0.228} & \textbf{60.5\%} & \textbf{88.4\%} & 96.2\% \\ \midrule
UD & \textbf{0.134} & \textbf{0.119} & \textbf{0.571} & \textbf{0.208} & \textbf{66.4\%} & \textbf{90.8\%} & \textbf{96.8\%} \\
w/o CC & 0.138 & 0.136 & 0.650 & 0.224 & 61.2\% & 88.9\% & 96.3\% \\ \bottomrule
\end{tabular}
\vspace{-0.15in}
\end{table}

\textbf{Spherical Attention:}
We conduct two experiments to assess the gains associated to the spherical attention maps by re-training UD and LR without their respective attention masks $A_{UD}$ and $A_{LR}$.
The results are presented in Table \ref{tab:attention_ablation} where an interesting outcome is apparent.
Their effect on LR is significant while for UD it remains questionable as it very slightly hampers performance.
ERP distortions are more prevalent in LR and stabilizing the loss during training by reducing their effect, is very important.
On the contrary, vertical distortions are gracefully handled by DIBR, therefore rendering the attention insignificant.

\begin{table}[!htbp]
\centering
\caption{Spherical Attention Ablation Results}
\label{tab:attention_ablation}
\small\addtolength{\tabcolsep}{-4pt}
\begin{tabular}{@{}cccccccc@{}}
\toprule
            & \cellcolor[HTML]{DAE8FC} \begin{tabular}[c]{@{}c@{}}Abs\\ Rel\end{tabular} & \cellcolor[HTML]{DAE8FC} \begin{tabular}[c]{@{}c@{}}Sq\\ Rel\end{tabular} & \cellcolor[HTML]{DAE8FC} RMSE           & \multicolumn{1}{r}{\cellcolor[HTML]{DAE8FC}RMSLE} & $\cellcolor[HTML]{CBCEFB} \delta_1$              & $\cellcolor[HTML]{CBCEFB} \delta_2$              & $\cellcolor[HTML]{CBCEFB} \delta_3$              \\ \midrule
LR          & \textbf{0.143}                                    & \textbf{0.129}                                   & \textbf{0.639} & \textbf{0.230}            & \textbf{58.1\%} & \textbf{88.2\%} & \textbf{96.5\%} \\
w/o $A_{LR}$ & 0.269                                             & 0.295                                            & 0.824          & 0.324                     & 56.7\%          & 84.4\%          & 93.2\%          \\ \midrule
UD          & 0.134                                             & 0.119                                            & 0.571          & 0.208                     & \textbf{66.4\%} & \textbf{90.8\%} & \textbf{96.8\%} \\
w/o $A_{UD}$ & \textbf{0.132}                                    & \textbf{0.116}                                   & \textbf{0.566} & \textbf{0.205}            & 66.2\%          & 90.1\%          & 96.0\%          \\ \bottomrule
\end{tabular}
\vspace{-0.1in}
\end{table}
\section{Discussion}
Spherical view synthesis is a relatively unexplored supervision scheme, mainly due to the lack of data and the challenges that it entails.
We have presented a learning scheme under which self-supervised \360 depth estimation is possible addressing the challenges mainly related to the distortions that ERP introduces.
Our work is the first to train a horizontal baseline \360 self-supervised model and to achieve this, besides introducing the full \360 disparity model, a more robust \360 view synthesis was required.
The DIBR splatting scheme, in combination with spherical attention, manage to overcome the inconsistent supervision that traditional inverse warping approaches suffer from.
Nonetheless, vertical stereo setups are offering higher quality models, further improved by CoordConvs, but as current research focuses on utilizing videos for learning depth estimation, the challenges that horizontal disparity comes with, as well as the full spherical disparity model, are very relevant. 
Finally, an unsurprising open question is raised with respect to the performance deviation of self-supervised and fully supervised models.
Is self-supervision the direction to pursue, or are other approaches like higher quality data acquisition, or synthetic data and domain adaptation, perhaps, better alternatives?

\textbf{Acknowledgements:}
We acknowledge HW support by Nvidia and financial support by the H2020 EC project Hyper360 (GA 761934).

{\small
\bibliographystyle{ieee}
\bibliography{egbib}
}

\end{document}